\documentclass[11pt,a4paper]{article}
\usepackage[hyperref]{acl2019}
\usepackage{times}
\usepackage{latexsym}
\usepackage{graphicx}
\usepackage{blindtext}
\usepackage{amsmath}
\usepackage{mathtools}%http://ctan.org/pkg/mathtools
\usepackage{relsize}
\usepackage{dsfont}
\usepackage{nicefrac}
\usepackage[multiple]{footmisc}
\usepackage{lipsum}

\usepackage{url}
\usepackage{dblfloatfix}    % To enable figures at the bottom of page
\usepackage{subcaption}
\usepackage[disable]{todonotes}
\usepackage{soul,color}
\usepackage[normalem]{ulem}
\usepackage{titlesec}
\usepackage{courier}
\usepackage{ulem}

\newcommand*\mean[1]{\bar{#1}}

\aclfinalcopy % Uncomment this line for the final submission
\title{Analyzing the Structure of Attention in a Transformer Language Model}

\author{Jesse Vig \\
  Palo Alto Research Center \\
    Machine Learning and \\ Data Science Group \\
  Interaction and Analytics Lab \\
  Palo Alto, CA, USA  \\
  \texttt{jesse.vig@parc.com} \\\And
  Yonatan Belinkov \\
  Harvard John A. Paulson School of \\  Engineering and Applied Sciences and   \\ 
  MIT Computer Science and \\ Artificial Intelligence Laboratory \\
  Cambridge, MA, USA \\
  \texttt{belinkov@seas.harvard.edu} \\}

\date{}

\begin{document}
\maketitle
\begin{abstract}
The Transformer is a fully attention-based alternative to recurrent networks that has achieved state-of-the-art results across a range of NLP tasks. In this paper, we analyze the structure of attention in a Transformer language model, the GPT-2 small pretrained model. We visualize attention for individual instances and analyze the interaction between attention and syntax over a large corpus. We find that attention targets different parts of speech at different layer depths within the model, and that attention aligns with dependency relations most strongly in the middle layers. We also find that the deepest layers of the model capture the most distant relationships. Finally, we extract exemplar sentences that reveal highly specific patterns targeted by particular attention heads.
\end{abstract}
\section{Introduction}

Contextual word representations have recently been used to achieve state-of-the-art performance across a range of language understanding tasks~\cite{peters-etal-2018-deep,Radford2018IL,Bert2018}. These representations are obtained by optimizing a language modeling (or similar) objective on large amounts of text. The underlying architecture may be recurrent, as in  ELMo~\cite{peters-etal-2018-deep}, or based on multi-head self-attention, as in OpenAI's GPT~\cite{Radford2018IL} and BERT~\cite{Bert2018}, which are based on the Transformer~\cite{transformer_arxiv}. Recently, the GPT-2 model~\cite{gpt2} outperformed other language models in a zero-shot setting, again based on self-attention. 

An advantage of using attention is that it can help interpret the model by showing how the model attends to different parts of the input~\citep{align_translate, belinkov:2018:tacl}.  Various tools have been developed to visualize attention in NLP models, ranging from attention matrix heatmaps~\citep{align_translate, neural_attention, rocktaschel2016reasoning} to bipartite graph representations~\citep{visual_interrogation, Lee2017, seq2seqvisv1}.  A visualization tool designed specifically for multi-head self-attention in the Transformer~\citep{JonesViz, tensor2tensor} was introduced in~\citet{transformer_arxiv}.

We extend the work of~\citet{JonesViz}, by visualizing attention in the Transformer at three levels of granularity: the attention-head level, the model level, and the neuron level. We also adapt the original encoder-decoder implementation to the decoder-only GPT-2 model, as well as the encoder-only BERT model.

In addition to visualizing attention for individual inputs to the model, we also analyze attention in aggregate over a large corpus to answer the following research questions:
\begin{itemize}
\item{Does attention align with syntactic dependency relations?}
\item{Which attention heads attend to which part-of-speech tags?}
\item{How does attention capture long-distance relationships versus short-distance ones?}
\end{itemize}
We apply our analysis to the GPT-2 small pretrained model. We find that attention follows dependency relations most strongly in the middle layers of the model, and that attention heads target particular parts of speech depending on layer depth. We also find that attention spans the greatest distance in the deepest layers, but varies significantly between heads. Finally, our method for extracting exemplar sentences yields many intuitive patterns. 

\section{Related Work}
\label{sec:related}
Recent work suggests that the Transformer  implicitly encodes syntactic information such as dependency parse trees~\cite{hewitt2019probe, raganato-tiedemann-2018-analysis}, anaphora~\cite{voita-etal-2018-context}, and subject-verb pairings~\cite{goldberg2019bert, Wolf2019GPT}. Other work has shown that RNNs also capture syntax, and that deeper layers in the model capture increasingly high-level constructs~\cite{blevins-2018-syntax}.

In contrast to past work that measure a model's syntactic knowledge through linguistic probing tasks, we directly compare the model's attention patterns to syntactic constructs such as dependency relations and part-of-speech tags.~\citet{raganato-tiedemann-2018-analysis} also evaluated dependency trees induced from attention weights in a Transformer, but in the context of encoder-decoder translation models.

\section{Transformer Architecture}
\label{sec:transformer}
\paragraph{Stacked Decoder:}
GPT-2 is a stacked decoder Transformer, which inputs a sequence of tokens and applies position and token embeddings followed by several decoder layers. Each layer applies multi-head self-attention (see below) in combination with a feedforward network, layer normalization, and residual connections. 
The GPT-2 small model has 12 layers and 12 heads.

\paragraph{Self-Attention:}
Given an input $x$, the self-attention mechanism assigns to each token $x_i$ a set of attention weights over the tokens in the input:
\begin{equation}
\text{Attn}(x_i) = (\alpha_{i, 1}(x), \alpha_{i, 2}(x), ..., \alpha_{i, i}(x))
\end{equation}
where $\alpha_{i, j}(x)$ is the attention that $x_i$ pays to $x_j$. The weights are positive and sum to one. Attention in GPT-2 is right-to-left, so $\alpha_{i, j}$ is defined only for $j \leq i$. In the multi-layer, multi-head setting, $\alpha$ is specific to a layer and head. 

The attention weights $\alpha_{i, j}(x)$ are computed from the scaled dot-product of the \textit{query vector} of $x_i$ and the \textit{key vector} of $x_j$, followed by a softmax operation. The attention weights are then used to produce a weighted sum of value vectors:

\begin{equation}
    \text{Attention}(Q, K, V) = \text{softmax}(\frac{QK^T}{\sqrt{d_k}})V
    \label{eq:attn_computation}
\end{equation}
using query matrix $Q$, key matrix $K$, and value matrix $V$, where $d_k$ is the dimension of $K$. In a multi-head setting, the queries, keys, and values are linearly projected $h$ times, and the attention operation is performed in parallel for each representation, with the results concatenated.

\section{Visualizing Individual Inputs}
In this section, we present three visualizations of attention in the Transformer model: the attention-head view, the model view, and the neuron view. Source code and Jupyter notebooks are available at \url{https://github.com/jessevig/bertviz}, and a video demonstration can be found at \url{https://vimeo.com/339574955}. A more detailed discussion of the tool is provided in~\citet{vig-2019-acl-bertviz}.

\subsection{Attention-head View}
\begin{figure}[h]
    \centering
    \vspace{.3em}
    \includegraphics[width=1\linewidth]{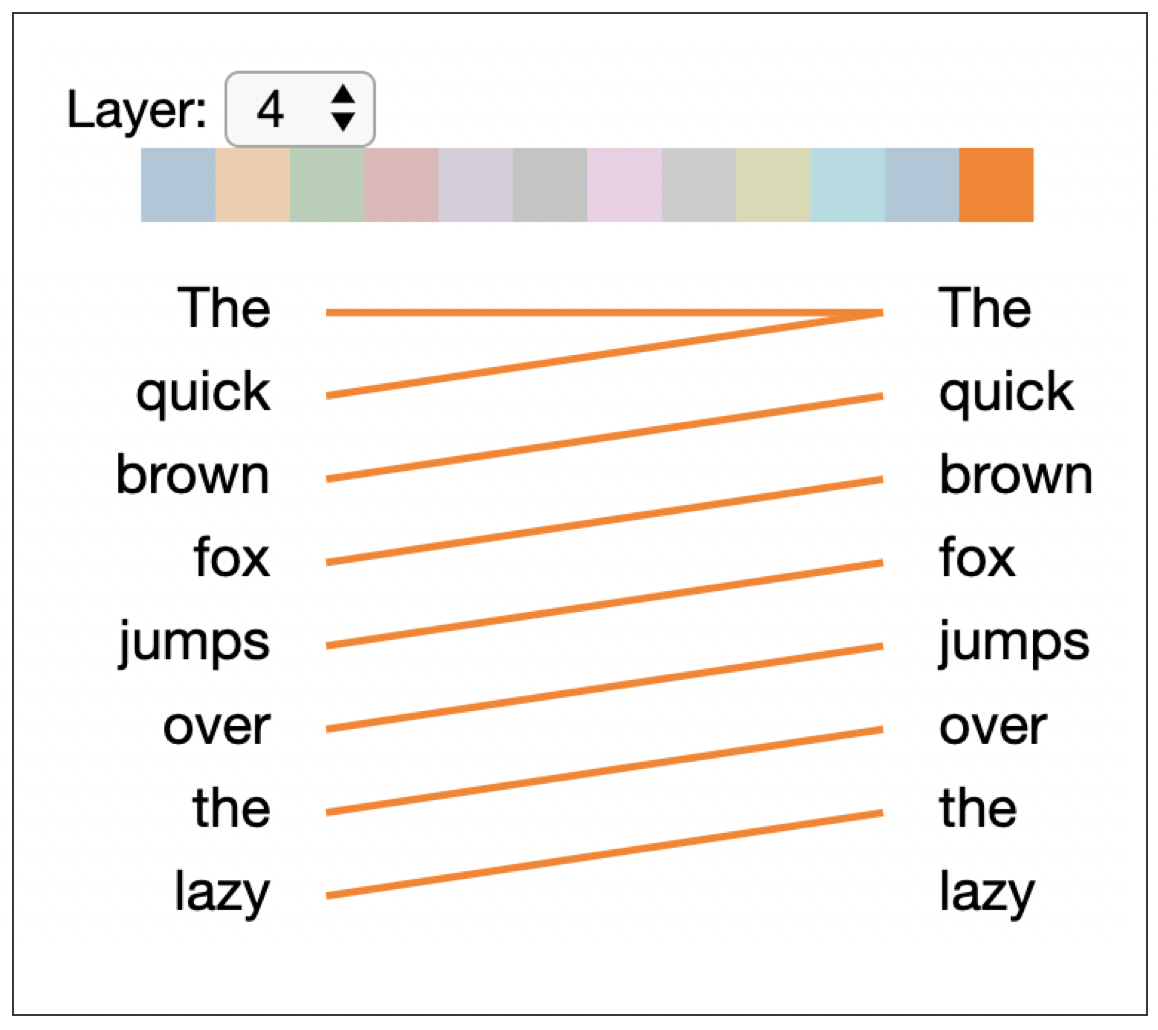}
    \caption{Attention-head view of GPT-2 for layer 4, head 11, which focuses attention on previous token.}
    \label{fig:head_view_3}
\end{figure}

The \textit{attention-head view} (Figure~\ref{fig:head_view_3}) visualizes attention for one or more heads in a model layer. Self-attention is depicted as lines connecting the attending tokens (left) with the tokens being attended to (right). Colors identify the head(s), and line weight reflects the attention weight. This view closely follows the design of~\citet{JonesViz}, but has been adapted to the GPT-2 model (shown in the figure) and BERT model (not shown).

This view helps focus on the role of specific attention heads. For instance, in the shown example, the chosen attention head attends primarily to the previous token position.

\begin{figure}[t]
    \includegraphics[width=.9\linewidth]{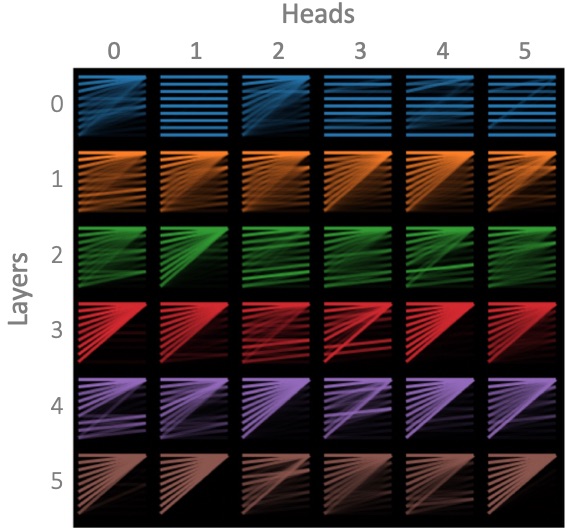}
    \caption{Model view of GPT-2, for same input as in Figure~\ref{fig:head_view_3} (excludes layers 6--11 and heads 6--11).}
    \label{fig:model_view}
\end{figure}

\begin{figure*}[t]
    \includegraphics[width=1\linewidth]{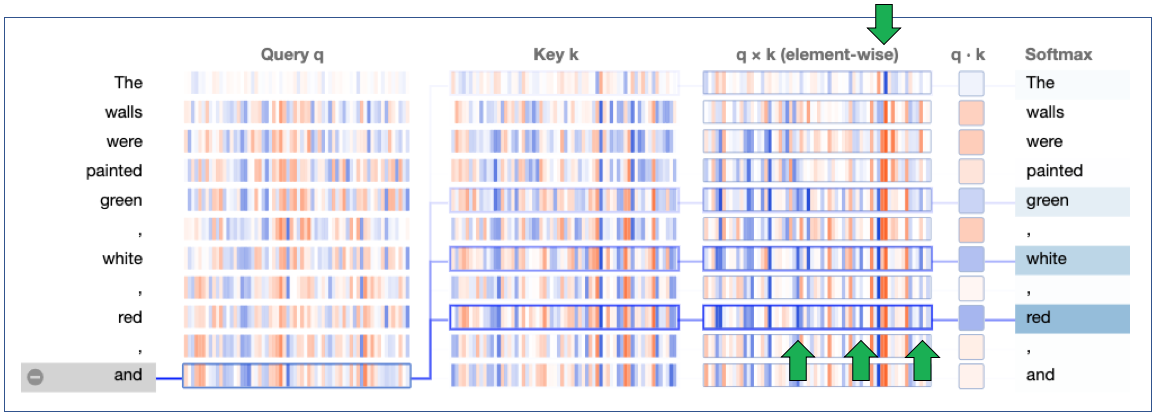}
    \vspace{-1.7em}
    \caption{Neuron view for layer 8, head 6, which targets items in lists. Positive and negative values are colored blue and orange, respectively, and color saturation indicates magnitude. This view traces the computation of attention (Section \ref{sec:transformer}) from the selected token on the left to each of the tokens on the right. Connecting lines are weighted based on attention between the respective tokens. The arrows (not in visualization) identify the neurons that most noticeably contribute to this attention pattern: the lower arrows point to neurons that contribute to attention towards list items, while the upper arrow identifies a neuron that helps focus attention on the first token in the sequence.}
    \vspace{-.2em}
    \label{fig:neuron_view}
\end{figure*}

\subsection{Model View}
The \textit{model view} (Figure~\ref{fig:model_view}) visualizes attention across all of the model's layers and heads for a particular input. Attention heads are presented in tabular form, with rows representing layers and columns representing heads.  Each head is shown in a thumbnail form that conveys the coarse shape of the attention pattern, following the \textit{small multiples} design pattern~\citep{Tufte1990}. Users may also click on any head to enlarge it and see the tokens. 

This view facilitates the detection of coarse-grained differences between heads. For example, several heads in layer 0 share a horizontal-stripe pattern, indicating that tokens attend to the current position. Other heads have a triangular pattern, showing that they attend to the first token. In the deeper layers, some heads display a small number of highly defined lines, indicating that they are targeting specific relationships between tokens. 

\subsection{Neuron View}

The \textit{neuron view} (Figure~\ref{fig:neuron_view}) visualizes how individual neurons interact to produce attention. This view displays the queries and keys for each token, and demonstrates how attention is computed from the scaled dot product of these vectors. The element-wise product shows how specific neurons influence the dot product and hence attention.
 
Whereas the attention-head view and the model view show \textit{what} attention patterns the model learns, the neuron view shows \textit{how} the model forms these patterns. For example, it can help identify neurons responsible for specific attention patterns, as illustrated in Figure~\ref{fig:neuron_view}.

\section{Analyzing Attention in Aggregate}

In this section we explore the aggregate properties of attention across an entire corpus. We examine how attention interacts with syntax, and we compare long-distance versus short-distance relationships. We also extract exemplar sentences that reveal patterns targeted by each attention head.

\subsection{Methods}

\subsubsection{Part-of-Speech Tags}
\label{sec:method_pos}
Past work suggests that attention heads in the Transformer may specialize in particular linguistic phenomena~\cite{transformer_arxiv, raganato-tiedemann-2018-analysis, vig-2019-acl-bertviz}. We explore whether individual attention heads in GPT-2 target particular parts of speech. Specifically, we measure the proportion of total attention from a given head that focuses on tokens with a given part-of-speech tag, aggregated over a corpus: 
\vspace{-.5em}
\begin{equation}
\text{P}_{\alpha}(tag) = 
    \dfrac{
   \sum\limits_{x \in X} \sum\limits_{i=1}^{|x|}\sum\limits_{j=1}^i
    \alpha_{i,j}(x) {\cdot} \mathds{1}_{\text{pos}(x_j) = tag} 
    }
    {
   \sum\limits_{x \in X} \sum\limits_{i=1}^{|x|}\sum\limits_{j=1}^i
    \alpha_{i,j}(x)
    }
\label{eq:pos}
\end{equation}
\noindent
where $tag$ is a part-of-speech tag, e.g., \textit{NOUN}, $x$ is a sentence from the corpus $X$, $\alpha_{i, j}$ is the attention from $x_i$ to $x_j$ for the given head (see Section \ref{sec:transformer}),   and $\text{pos}(x_j)$ is the part-of-speech tag of $x_j$. We also compute the share of attention directed \textit{from} each part of speech in a similar fashion.
\subsubsection{Dependency Relations}
\label{sec:method_dep}

Recent work shows that Transformers and recurrent models encode dependency relations~\citep{hewitt2019probe, raganato-tiedemann-2018-analysis, liu:2019:NAACL}. However, different models capture dependency relations at different layer depths. In a Transformer model, the middle layers were most predictive of dependencies~\cite{liu:2019:NAACL, nlp-pipeline}. Recurrent models were found to encode dependencies in lower layers for language models
~\cite{liu:2019:NAACL} and in deeper layers for translation models~\cite{belinkov-2018-thesis}.  

We analyze how attention aligns with dependency relations in GPT-2 by computing the proportion of attention that connects tokens that are also in a dependency relation with one another. We refer to this metric as \textit{dependency alignment}: 

\begin{equation}
\text{DepAl}_{\alpha} = 
    \dfrac{
   \sum\limits_{x \in X} \sum\limits_{i=1}^{|x|}\sum\limits_{j=1}^i
    \alpha_{i,j}(x)  dep(x_i, x_j)
    }
    {
   \sum\limits_{x \in X} \sum\limits_{i=1}^{|x|}\sum\limits_{j=1}^i
    \alpha_{i,j}(x)
    }
\label{eq:dep}
\end{equation}

\noindent
where $dep(x_i, x_j)$ is an indicator function that returns 1 if $x_i$ and $x_j$ are in a dependency relation and 0 otherwise. We run this analysis under three alternate formulations of dependency: (1) the attending token ($x_i$) is the parent in the dependency relation, (2) the token receiving attention ($x_j$) is the parent, and (3) either token is the parent.

We hypothesized that heads that focus attention based on position---for example, the head in Figure~\ref{fig:head_view_3} that focuses on the previous token---would not align well with dependency relations, since they do not consider the content of the text.  To distinguish between content-dependent and content-independent (position-based) heads, we define \textit{attention variability}, which measures how attention varies over different inputs; high variability would suggest a content-dependent head, while low variability would indicate a content-independent head:

\vspace{-.6em}
\begin{equation}
\text{Variability}_{\alpha} = 
    \dfrac{
   \sum\limits_{x \in X} \sum\limits_{i=1}^{|x|}\sum\limits_{j=1}^i
    |\alpha_{i,j}(x)-\mean{\alpha}_{i,j}| 
    }
    { 2\cdot\sum\limits_{x \in X} \sum\limits_{i=1}^{|x|}\sum\limits_{j=1}^i
    \alpha_{i,j}(x)
    }
\label{eq:var}
\end{equation}
where $\mean{\alpha}_{i,j}$ is the mean of $\alpha_{i,j}(x)$ over all $x \in X$.

\begin{figure*}[tp]
\centering
    \includegraphics[width=.99\linewidth]{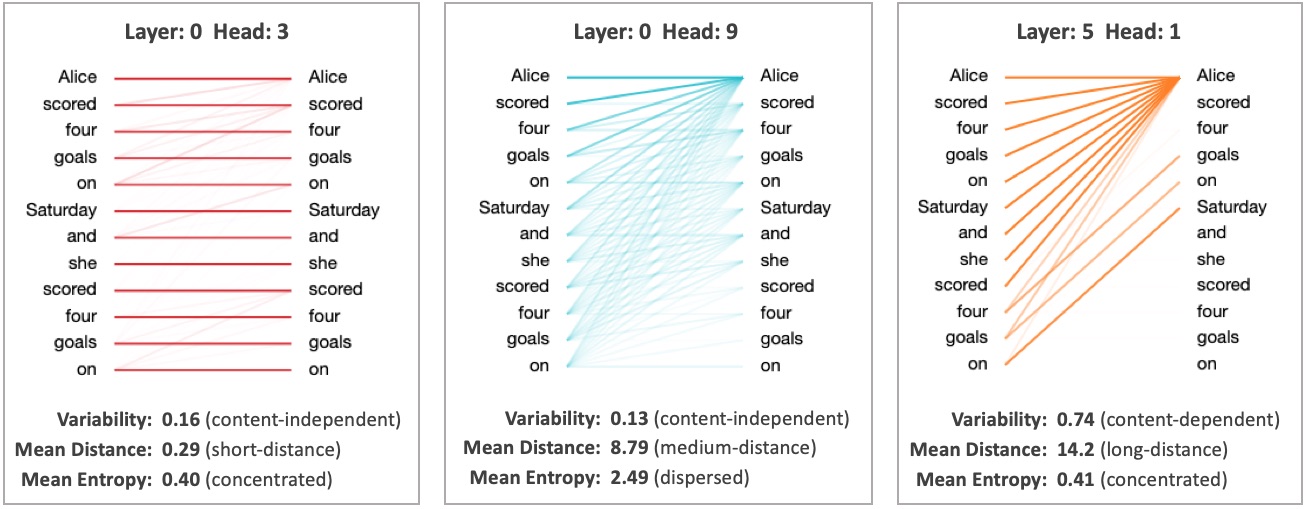}
    \caption{Attention heads in GPT-2 visualized for an example input sentence, along with aggregate metrics computed from all sentences in the corpus. Note that the average sentence length in the corpus is 27.7 tokens. \textbf{Left:} Focuses attention primarily on current token position. \textbf{Center:} Disperses attention roughly evenly across all previous tokens. \textbf{Right:} Focuses on words in repeated phrases. }
    \label{fig:examples_combined_2}
\end{figure*}

$\text{Variability}_\alpha$ represents the mean absolute deviation\footnote{We considered using variance to measure attention variability; however, attention is  sparse for many attention heads after filtering first-token attention (see Section \ref{sub:preprocessing}), resulting in a very low variance (due to $\alpha_{i,j}(x) \approx 0$ and $\mean{\alpha}_{i,j} \approx 0$) for many content-sensitive attention heads. We did not use a probability distance measure, as attention values do not sum to one due to filtering first-token attention. } of $\alpha$ over $X$, scaled to the $\lbrack 0,1 \rbrack$ interval.\footnote{The upper bound is 1 because the denominator is an upper bound on the numerator.}\footnote{When computing variability, we only include the first $N$ tokens ($N$=10) of each $x \in X$ to ensure a sufficient amount of data at each position $i$. The positional patterns appeared to be consistent across the entire sequence.} Variability scores for three example attention heads are shown in Figure~\ref{fig:examples_combined_2}.

\subsubsection{Attention Distance}  

Past work suggests that deeper layers in NLP models capture longer-distance relationships than lower layers~\cite{belinkov-2018-thesis, raganato-tiedemann-2018-analysis}. We test this hypothesis on GPT-2 by measuring the mean distance (in number of tokens) spanned by attention for each head. Specifically, we compute the average distance between token pairs in all sentences in the corpus, weighted by the attention between the tokens:

\begin{equation}
\mean{D}_{\alpha} =
    \dfrac{
   \sum\limits_{x \in X} \sum\limits_{i=1}^{|x|}\sum\limits_{j=1}^i
    \alpha_{i,j}(x) \cdot (i - j) 
    }
    {
   \sum\limits_{x \in X} \sum\limits_{i=1}^{|x|}\sum\limits_{j=1}^i
    \alpha_{i,j}(x)
    }
\label{eq:dist}
\end{equation}

We also explore whether heads with more dispersed attention patterns (Figure~\ref{fig:examples_combined_2}, center) tend to capture more distant relationships. We measure attention dispersion based on the entropy\footnote{When computing entropy, we exclude attention to the first (null) token (see Section \ref{sub:preprocessing}) and renormalize the remaining weights. We exclude tokens that focus over 90\% of attention to the first token, to avoid a disproportionate influence from the remaining attention from these tokens.} of the attention distribution~\cite{ghader-monz-2017-attention}: 

\begin{equation}
\text{Entropy}_{\alpha}(x_i) = - \sum_{j = 1}^{i} \alpha_{i,j}(x)\text{log}(\alpha_{i,j}(x))
\label{eq:entropy}
\end{equation}

Figure~\ref{fig:examples_combined_2} shows the mean distance and entropy values for three example attention heads.

\begin{figure}
    \centering
    \includegraphics[width=.85\linewidth]{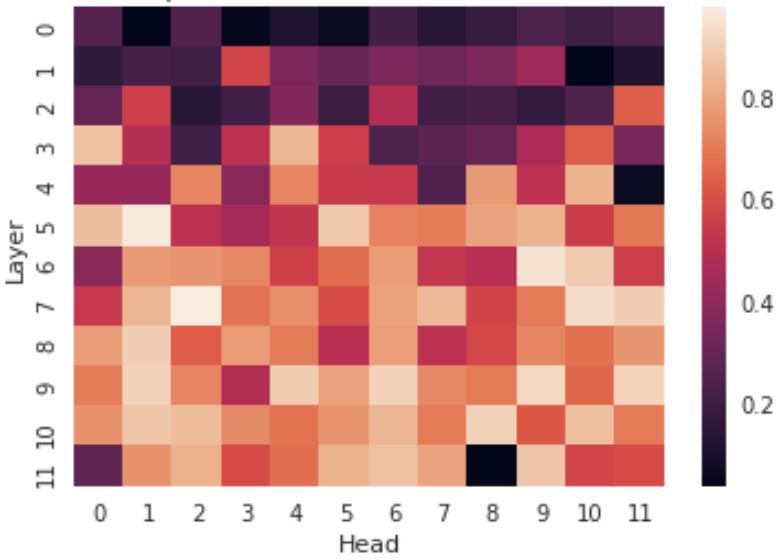}
    \caption{Proportion of attention focused on first token, broken out by layer and head.}
    \label{fig:first_token_attn}
\end{figure}

\subsection{Experimental Setup}

\subsubsection{Dataset}
We focused our analysis on text from English Wikipedia,
which was not included in the training set for GPT-2. We first extracted 10,000 articles, and then sampled 100,000 sentences from these articles. For the qualitative analysis described later, we used the full dataset; for the quantitative analysis, we used a subset of 10,000 sentences.

\begin{figure*}[tp]
    \centering
    \includegraphics[width=1\linewidth]{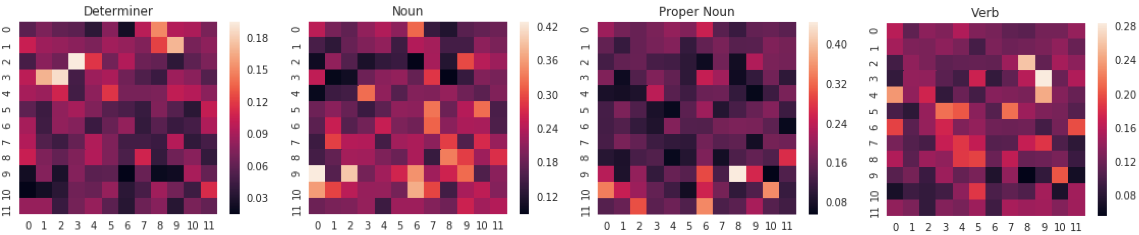}
    \vspace{-1em}
    \caption{Each heatmap shows the proportion of total attention directed \textit{to} the given part of speech, broken out by layer (vertical axis) and head (horizontal axis).  Scales vary by tag. Results for all tags available in appendix.}
    \vspace{1em}
    \label{fig:to_pos_10000}
    \includegraphics[width=01\linewidth]{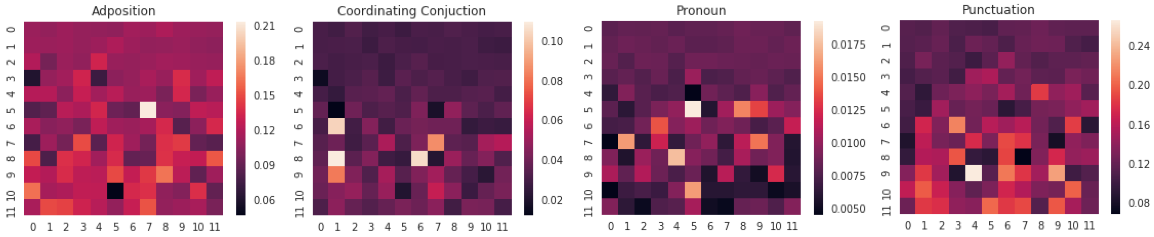}
\vspace{-1em}
  \caption{Each heatmap shows the proportion of total attention that originates \textit{from} the given part of speech, broken out by layer (vertical axis) and head (horizontal axis).  Scales vary by tag. Results for all tags available in appendix.}
    \label{fig:from_pos_10000}
\end{figure*}

\subsubsection{Tools}
We computed attention weights using the \texttt{pytorch-pretrained-BERT}\footnote{ \url{https://github.com/huggingface/pytorch-pretrained-BERT}} implementation of the GPT-2 small model. 
We extracted syntactic features using spaCy~\citep{spacy2} and mapped the features from the  spaCy-generated tokens to the corresponding tokens from the GPT-2 tokenizer.\footnote{In cases where the GPT-2 tokenizer split a word into multiple pieces, we assigned the features to all word pieces.}

\subsubsection{Filtering Null Attention}
\label{sub:preprocessing}
We excluded attention focused on the first token of each sentence from the analysis because it was not informative; other tokens appeared to focus on this token by default when no relevant tokens were found elsewhere in the sequence. On average, 57\% of attention was directed to the first token. Some heads focused over 97\% of attention to this token on average (Figure~\ref{fig:first_token_attn}), which is consistent with recent work showing that individual attention heads may have little impact on overall model performance~\cite{voita-etal-2019-pruning-heads, michel-etal-2019-pruning-heads}.
We refer to the attention directed to the first token as \textit{null attention}.

\begin{figure*}[t]
    \includegraphics[width=1\linewidth]{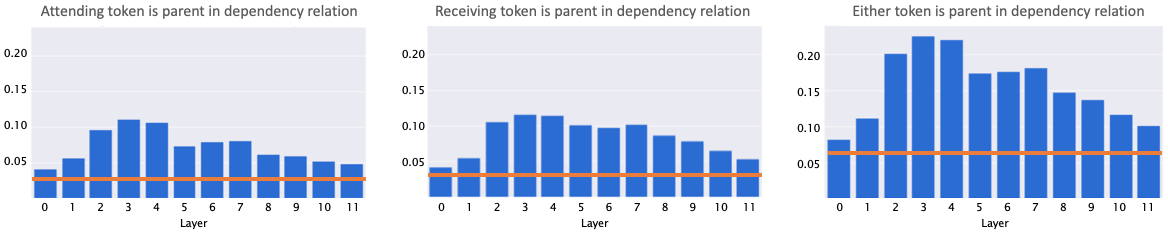}
    \vspace{-1.7em}
    \caption{Proportion of attention that is aligned with dependency relations, aggregated by layer. The orange line shows the baseline proportion of token pairs that share a dependency relationship, independent of attention.}
    \label{fig:dep_relation}
% \end{figure*}
    \vspace{1em}
% \begin{figure*}[t]
    \centering
    \includegraphics[width=\linewidth]{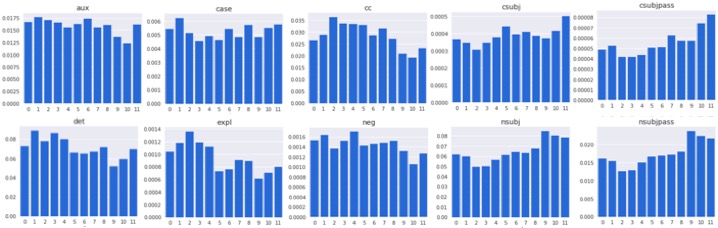}
        \vspace{-1.7em}
    \caption{Proportion of attention directed to various dependency types, broken out by layer.}
    % \vspace{-.5em}
    \label{fig:dep_class_to}
\end{figure*}

\begin{figure}[t]
    \centering
    \includegraphics[width=0.88\linewidth]{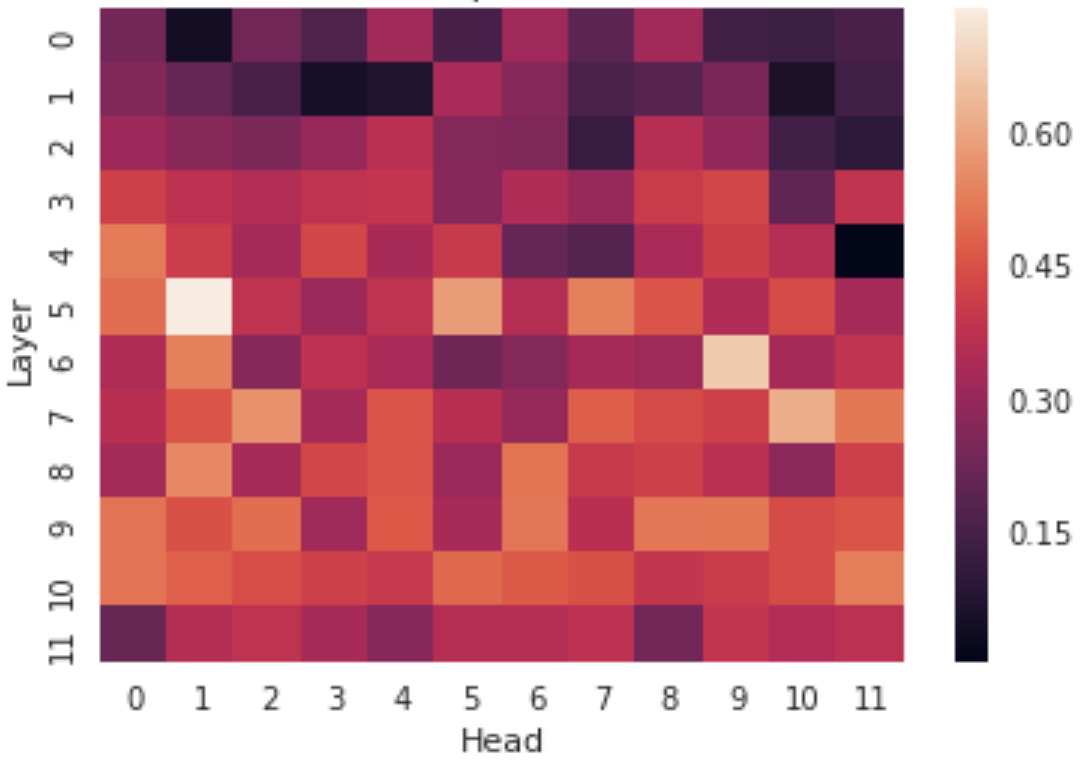}
    \caption{Attention variability by layer / head.  High-values indicate content-dependent heads, and low values indicate content-independent (position-based) heads.}
    \label{fig:attention_variance}
\end{figure}

\subsection{Results}

\subsubsection{Part-of-Speech Tags}
\label{sec:results_pos}

Figure~\ref{fig:to_pos_10000} shows the share of attention directed \textit{to} various part-of-speech tags (Eq. \ref{eq:pos}) broken out by layer and head. Most tags are disproportionately targeted by one or more attention heads. For example, nouns receive 43\% of attention in layer 9, head 0, compared to a mean of 21\% over all heads.
% Verbs attract 28.4\% of attention in layer 3, head 9, compared to an average of 13.3\%.
For 13 of 16 tags, a head exists with an attention share more than double the mean for the tag.

The attention heads that focus on a particular tag tend to cluster by layer depth. For example, the top five heads targeting proper nouns are all in the last three layers of the model. This may be due to several attention heads in the deeper layers focusing on named entities (see Section \ref{sub:qualitative}), which may require the broader context available in the deeper layers.
In contrast, the top five heads targeting determiners---a lower-level construct---are all in the first four layers of the model. This is consistent with previous findings showing that deeper layers focus on higher-level properties~\cite{blevins-2018-syntax, belinkov-2018-thesis}. 

 Figure~\ref{fig:from_pos_10000} shows the proportion of attention directed \textit{from} various parts of speech. The values appear to be roughly uniform in the initial layers of the model. The reason is that the heads in these layers pay little attention to the first (null) token (Figure~\ref{fig:first_token_attn}), and therefore the remaining (non-null) attention weights sum to a value close to one. Thus, the net weight for each token in the weighted sum (Section \ref{sec:method_pos}) is close to one, and the proportion reduces to the frequency of the part of speech in the corpus.
 
Beyond the initial layers, attention heads specialize in focusing attention from particular part-of-speech tags. However, the effect is less pronounced compared to the tags receiving attention; for 7 out of 16 tags, there is a head that focuses attention from that tag with a frequency more than double the tag average. Many of these specialized heads also cluster by layer. For example, the top ten heads for focusing attention from punctuation are all in the last six layers.

\begin{figure*}[tp]
    \centering
    \includegraphics[width=.95\linewidth]{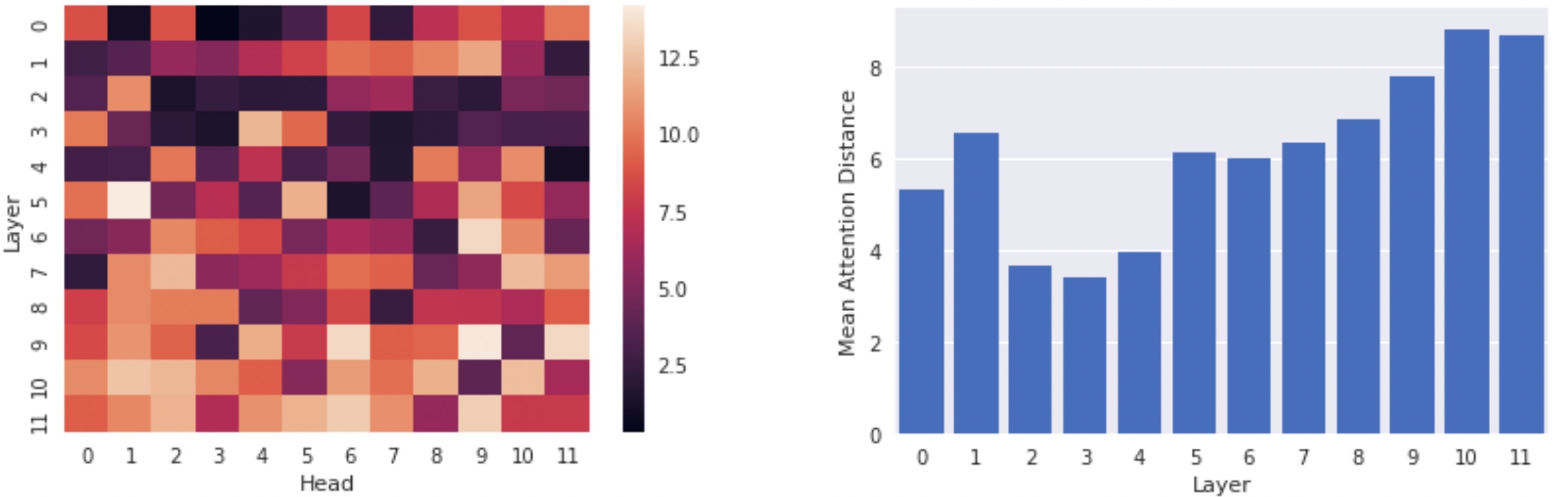}
        \vspace{-.1em}
    \caption{Mean attention distance by layer / head (left), and by layer (right).}
    \vspace{-.5em}
    \label{fig:token_distance_combined}
\end{figure*}

\begin{figure}[h]
    \centering
    \includegraphics[width=0.85\linewidth]{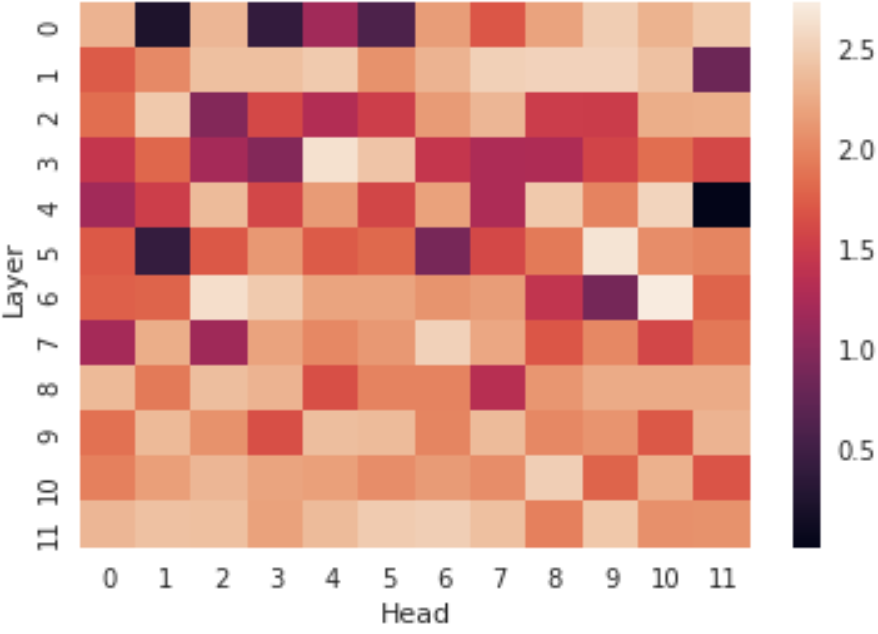}
    \vspace{-.1em}
    \caption{Mean attention entropy by layer / head. Higher values indicate more diffuse attention.}
        \vspace{-.5em}
    \label{fig:entropy}
\end{figure}

\subsubsection{Dependency Relations}

Figure~\ref{fig:dep_relation} shows the dependency alignment scores (Eq. \ref{eq:dep}) broken out by layer.
Attention aligns with dependency relations  most strongly in the middle layers, consistent with recent syntactic probing analyses~\cite{liu:2019:NAACL, nlp-pipeline}. 

One possible explanation for the low alignment in the initial layers is that many heads in these layers focus attention based on position rather than content, according to the attention variability (Eq. \ref{eq:var}) results in Figure~\ref{fig:attention_variance}. Figure~\ref{fig:examples_combined_2} (left and center) shows two examples of  position-focused heads from layer 0 that have relatively low dependency alignment\footnotemark[7] (0.04 and 0.10, respectively); the first head focuses attention primarily on the current token position (which cannot be in a dependency relation with itself) and the second disperses attention roughly evenly, without regard to content. 

An interesting counterexample is layer 4, head 11 (Figure~\ref{fig:head_view_3}), which has the highest dependency alignment out of all the heads ($\text{DepAl}_{\alpha} = 0.42$)\footnote{Assuming relation may be in either direction.} but is also the most position-focused ($\text{Variability}_{\alpha} = 0.004$). This head  focuses attention on the previous token, which in our corpus has a 42\% chance of being in a dependency relation with the adjacent token. As we'll discuss in the next section, token distance is highly predictive of dependency relations.

One hypothesis for why attention diverges from dependency relations in the deeper layers is that several attention heads in these layers target very specific constructs (Tables \ref{tab:sentences1} and \ref{tab:sentences2}) as opposed to more general dependency relations.  The deepest layers also target longer-range relationships (see next section), whereas dependency relations span relatively short distances (3.89 tokens on average).  

We also analyzed the specific dependency types of tokens receiving attention (Figure~\ref{fig:dep_class_to}). Subjects (csubj, csubjpass, nsubj, nsubjpass) were targeted more in deeper layers, while auxiliaries (aux), conjunctions (cc), determiners (det), expletives (expl), and negations (neg) were targeted more in lower layers, consistent with previous findings~\cite{belinkov-2018-thesis}. For some other dependency types, the interpretations were less clear. 

\subsubsection{Attention Distance}
We found that attention distance (Eq. \ref{eq:dist}) is greatest in the deepest layers (Figure~\ref{fig:token_distance_combined}, right), confirming that these layers capture longer-distance relationships. Attention distance varies greatly across heads ($SD = 3.6$), even when the heads are in the same layer, due to the wide variation in attention structures (e.g., Figure~\ref{fig:examples_combined_2} left and center).

We also explored the relationship between attention distance and attention entropy (Eq. \ref{eq:entropy}), which measures how diffuse an attention pattern is. Overall, we found a moderate correlation ($r=0.61$, $p < 0.001$) between the two. As Figure~\ref{fig:entropy} shows, many heads in layers 0 and 1 have high entropy (e.g., Figure~\ref{fig:examples_combined_2}, center), which may explain why these layers have a higher attention distance compared to layers 2--4. 

One counterexample is layer 5, head 1 (Figure~\ref{fig:examples_combined_2}, right), which has the highest mean attention distance of any head (14.2), and one of the lowest mean entropy scores (0.41).  This head concentrates attention on individual words in repeated phrases, which often occur far apart from one another.

We also explored how attention distance relates to dependency alignment. Across all heads, we found a  negative correlation between the two quantities ($r=-0.73, p <0.001$). This is consistent with the fact that the probability of two tokens sharing a dependency relation decreases as the distance between them increases\footnote{This is true up to a distance of 18 tokens; 99.8\% of dependency relations occur within this distance.}; for example, the probability of being in a dependency relation is 0.42 for adjacent tokens, 0.07 for tokens at a distance of 5, and 0.02 for tokens at a distance of 10. The layers (2--4) in which attention spanned the shortest distance also had the highest dependency alignment. 

\setlength{\abovecaptionskip}{4pt plus 3pt minus 2pt}
\setul{2.6pt}{.8pt}
\bgroup
\def\arraystretch{1.13}%  1 is the default, change whatever you need

\subsection{Qualitative Analysis}
\label{sub:qualitative}
To get a sense of the lexical patterns targeted by each attention head, we extracted exemplar sentences that most strongly induced attention in that head. Specifically, we ranked sentences by the maximum token-to-token attention weight within each sentence. Results for three attention heads are shown in Tables \ref{tab:sentences1}--\ref{tab:sentences3}.
We found other attention heads that detected entities (people, places, dates), passive verbs, acronyms, nicknames, paired punctuation, and other syntactic and semantic properties. Most heads captured multiple types of patterns.

\begin{table*}[t]
  \centering
  \begin{tabular}{|p{1cm}|p{14cm}|}
\hline
\textbf{Rank} & \textbf{Sentence} \\
\hline
1 & The  Australian  search  and  rescue  service  is  provided  by  Aus S AR ,  which  is  part  of  the  Australian  Maritime  \ul{\textit{Safety}}  Authority  ( \ul{\textbf{AM}} SA ).\\
\hline
2 & In  1925 ,  Bapt ists  worldwide  formed  the  Baptist  \ul{\textit{World}}  Alliance  ( \ul{\textbf{B}} WA ). \\
\hline
3 & The  Oak dale  D ump  is  listed  as  an  Environmental  Protection  Agency  Super fund  site  due  to  the  contamination  of  residential  drinking  water  wells  with  \ul{\textit{volatile}}  organic  compounds  \ul{\textbf{(}} V OC s )  and  heavy  metals . \\
\hline
\end{tabular}
  \caption{Exemplar sentences for layer 10, head 10, which focuses attention from acronyms to the associated phrase. The tokens with maximum attention are underlined; the attending token is bolded and the token receiving attention is italicized. It appears that attention is directed to the part of the phrase that would help the model choose the \textit{next} word piece in the acronym (after the token paying attention), reflecting the language modeling objective.}
  \vspace{.4em}
  \label{tab:sentences1}
\centering
\begin{tabular}{|p{1cm}|p{14cm}|}
\hline
\textbf{Rank} & \textbf{Sentence} \\
\hline
1 & After the  two  prototypes  were  completed ,  production  began  in  Mar iet \ul{\textit{ta}} \ul{\textbf{\large{,}}}  Georgia , ...\\
\hline
3 & The  fictional  character  Sam  Fisher  of  the  Spl inter  Cell  video  game  series  by  Ubisoft  was  born  in  Tow \ul{\textit{son}} \ul{\textbf{\large{,}}}  as  well  as  residing  in  a  town house ,  as  stated  in  the  novel izations ...\\
\hline
4 & Suicide  bombers  attack  three  hotels  in  Am \ul{\textit{man}} \ul{\textbf{\large{,}}}  Jordan ,  killing  at  least  60  people .\\
\hline
\end{tabular}
  \caption{Exemplar sentences for layer 11, head 2, which focuses attention from commas to the preceding place name (or the last word piece thereof). The likely purpose of this attention head is to help the model choose the related place name that would follow the comma, e.g. the country or state in which the city is located. } 
\vspace{.4em}
  \label{tab:sentences2}
  \centering
  \begin{tabular}{|p{1cm}|p{14cm}|}
\hline
\textbf{Rank} & \textbf{Sentence} \\
\hline
1 & With  the  United  States  isolation ist  and  Britain  stout ly  refusing  to  make  the  " continental  commitment "  to  defend  France  on  the  same  scale  as  in  World  War  I ,  the  \ul{\textit{prospects}}  of  Anglo - American  assistance  in  another  war  with  \ul{\textbf{Germany}}  appeared  to  be  doubtful ... \\
\hline
2 & The  show  did  receive  a  somewhat  favorable  review  from  noted  critic  Gilbert  Se ld es  in  the  December  15 ,  1962  TV  Guide :  " The  whole  \ul{\textit{notion}}  on  which  The  Beverly  Hill bill ies  is  \ul{\textbf{founded}} is  an  encouragement  to  ignorance ...\\
\hline
3 & he  Arch im edes  won  significant  market  share  in  the  education  markets  of  the  UK ,  Ireland ,  Australia  and  New  Zealand ;  the  \ul{\textit{success}} of  the  Arch im edes  in  British  \ul{\textbf{schools}}  was  due  partly  to  its  predecessor  the  BBC  Micro  and  later  to  the  Comput ers  for  Schools  scheme ...\\
\hline
\end{tabular}
  \caption{Exemplar sentences for layer 11, head 10 which focuses attention from the end of a noun phrase to the head noun. In the first sentence, for example, the head noun is \textit{prospects} and the remainder of the noun phrase is \textit{ of  Anglo - American  assistance  in  another  war  with  \textbf{Germany}}. The purpose of this attention pattern is likely to predict the word (typically a verb) that follows the noun phrase, as the head noun is a strong predictor of this.} 
  \vspace{-.1em}
  \label{tab:sentences3}
\end{table*}
\egroup

\section{Conclusion}
In this paper, we analyzed the structure of attention in the GPT-2 Transformer language model. We found that many attention heads specialize in particular part-of-speech tags and that different tags are targeted at different layer depths. We also found that the deepest layers capture the most distant relationships, and that attention aligns most strongly with dependency relations in the middle layers where attention distance is lowest.

Our qualitative analysis revealed that the structure of attention is closely tied to the  training objective; for GPT-2, which was trained using left-to-right language modeling, attention often focused on words most relevant to predicting the next token in the sequence. For future work, we would like to extend the analysis to other Transformer models such as BERT, which has a bidirectional architecture and is trained on both token-level and sentence-level tasks. 

Although the Wikipedia sentences used in our analysis cover a diverse range of topics, they all follow a similar encyclopedic format and style. Further study is needed to determine how attention patterns manifest in other types of content, such as dialog scripts or song lyrics. We would also like to analyze attention patterns in text much longer than a single sentence, especially for new Transformer variants such as the Transformer-XL~\cite{transformerxl} and Sparse Transformer~\cite{child-etal-2019-sparse-transformer}, which can handle very long contexts.

We believe that interpreting a model based on attention is complementary to linguistic probing approaches (Section \ref{sec:related}). While linguistic probing precisely quantifies the amount of information encoded in various components of the model, it requires training and evaluating a probing classifier. Analyzing attention is a simpler process that also produces human-interpretable descriptions of model behavior, though recent work casts doubt on its role in explaining individual predictions~\cite{jain-wallace-2019-attention-explanation}. The results of our analyses were often consistent with those from probing approaches. 
\vspace{-.1em}
\section{Acknowledgements}
Y.B.\ was supported by the Harvard Mind, Brain, and Behavior Initiative. 

\bibliography{main}
\bibliographystyle{acl_natbib}
\clearpage
\appendix
\renewcommand\thefigure{\thesection.\arabic{figure}}    
\section{Appendix}

\setcounter{figure}{0}

\begin{figure*}[tp]
    \centering
    \includegraphics[width=0.95\linewidth]{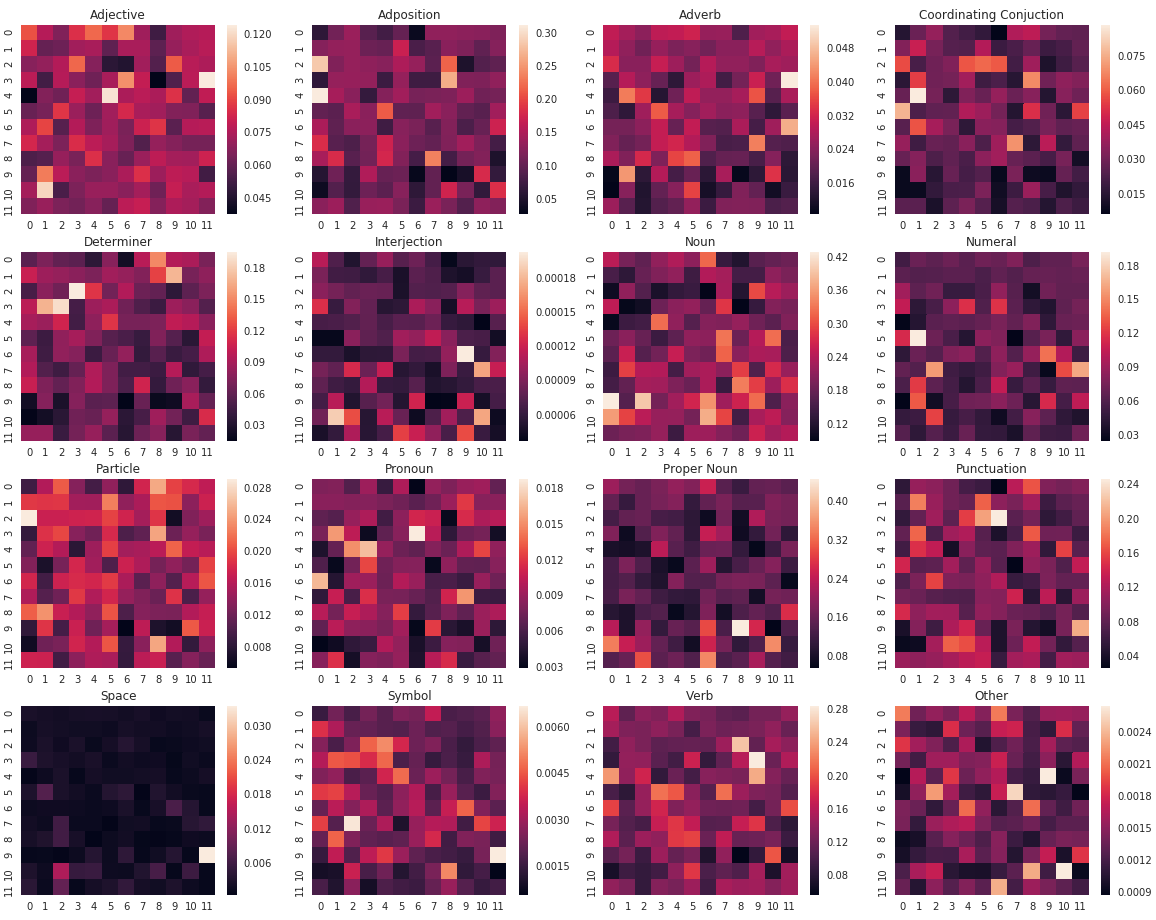}
    \caption{Each heatmap shows the proportion of total attention directed \textit{to} the given part of speech, broken out by layer (vertical axis) and head (horizontal axis).}
    \vspace*{4.2in}
    \label{fig:appendix1}
\end{figure*}

\begin{figure*}[t]
    \includegraphics[width=0.95\linewidth]{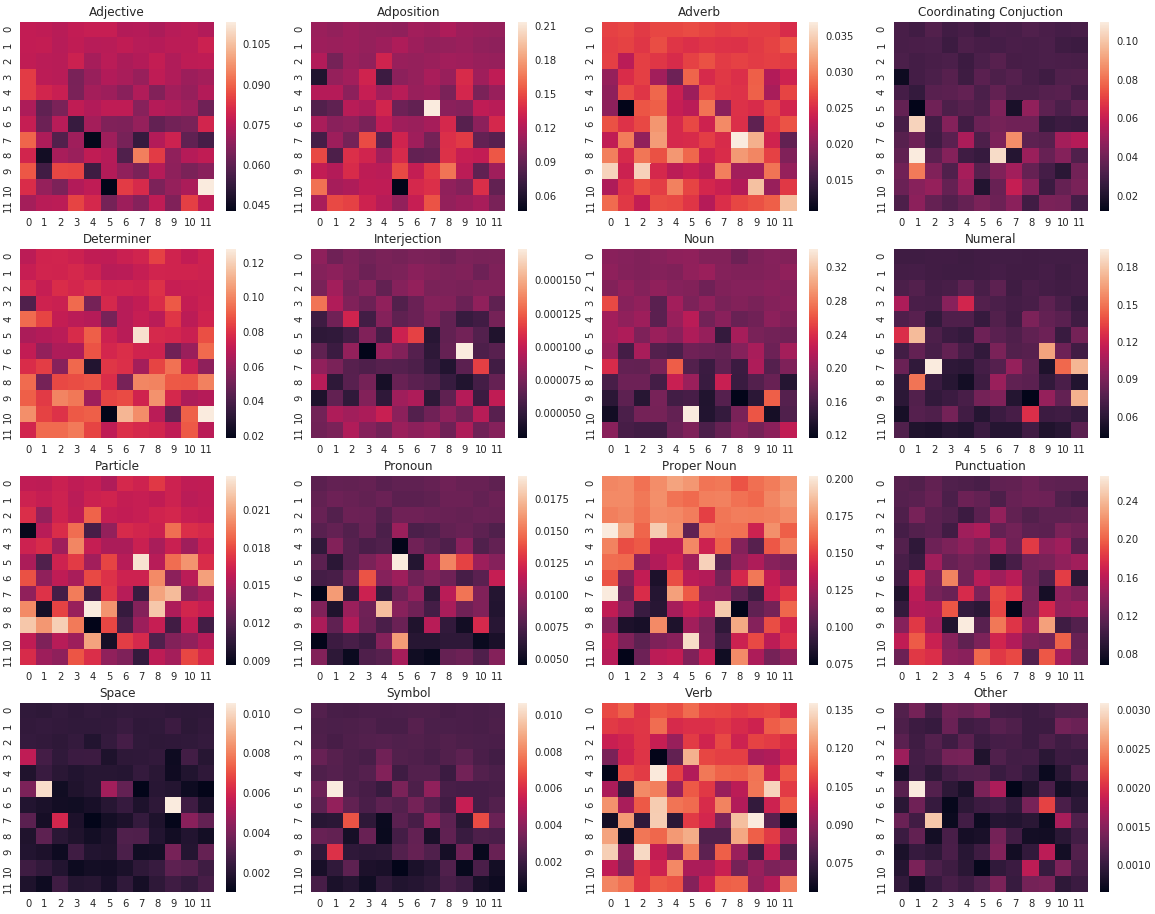}
    \caption{Each heatmap shows the proportion of attention originating \textit{from} the given part of speech, broken out by layer (vertical axis) and head (horizontal axis). }
    \vspace*{4.2in}
    \label{fig:appendix2}
\end{figure*}

Figures \ref{fig:appendix1} and \ref{fig:appendix2} shows the results from Figures \ref{fig:to_pos_10000} and \ref{fig:from_pos_10000} for the full set of part-of-speech tags.

% The appendix.
\clearpage

\end{document}